\documentclass{article}

\usepackage[final, nonatbib]{airov26}

\usepackage[utf8]{inputenc} 
\usepackage[T1]{fontenc}    
\usepackage{booktabs}       
\usepackage{amsfonts}       
\usepackage{nicefrac}       
\usepackage{microtype}      
\usepackage{xcolor}         
\usepackage{graphicx}
\usepackage{subcaption}
\usepackage{float}
\usepackage{siunitx}
\usepackage{comment}
\usepackage{listings}

\usepackage{svg}

\usepackage{acronym}
\acrodef{HRI}{Human Robot Interaction}
\acrodef{AI}{Artificial Intelligence}
\acrodef{ToF}{Time-of-Flight}
\acrodef{DoF}{Degrees-of-Freedom}
\acrodef{IMU}{Inertial Measurement Unit}
\acrodef{STIM}{Smart Transducer Interface Module}
\acrodef{TED}{Transducer Electronic Datasheet}
\acrodef{BLE}{Bluetooth Low Energy}
\acrodef{NCAP}{Network Capable Application Processor}
\acrodef{NFC}{Near-field Communication}
\acrodef{RGBD}{Color and Depth}
\acrodef{PCB}{Printed Circuit Board}
\acrodef{CDC}{Capacitance to Digital Converter}
\acrodef{GPU}{Graphics Processing Unit}
\acrodef{ROS}{Robot Operating System}
\acrodef{CNN}{Convolutional Neural Network}
\acrodef{RCNN}{Region-Based Convolutional Neural Network}
\acrodef{VLM}{Visual Language Model}
\acrodef{VLA}{Vision Language Action}
\acrodef{LLM}{Large Language Model}
\acrodef{MLLM}{Multimodal Large Language Model}
\acrodef{I2C}{Inter-Integrated Circuit}
\acrodef{CTSA}{Capacitive Tactile Sensor Array}
\acrodef{FOV}{field of view}
\acrodef{TSA}{Tactile Sensor Array}
\acrodef{SR}{success rate}
\acrodef{TPE}{thermoplastic elastomer}
\acrodef{UDP}{User Datagram Protocol}
\acrodef{FSN}{Flexible Silver Nanoparticle}
\acrodef{PDMS}{Polydimethylsiloxane}
\acrodef{TPV}{through-polymer-via}
\acrodef{RGB}{Color}
\acrodef{API}{Application Programming Interface}
\acrodef{ROC}{Receiver Operating Characteristic}
\acrodef{OCID}{Object Clutter Indoor Dataset}
\acrodef{OwlVit}{Vision Transformer for Open-World Localization}
\acrodef{COCO}{Common Objects in Context}
\acrodef{SAM}{Segment Anything}
\acrodef{YOLOv8}{You Only Look Once v8}
\acrodef{YOLOv4}{You Only Look Once v4}
\acrodef{CLIP}{Contrastive Language-Image Pre-Training}
\acrodef{PCA}{Principal Component Analysis}
\acrodef{RFID}{Radio Frequency Identification}
\acrodef{NIR}{Near Infrared}
\acrodef{VIS}{Visual Spectrometry}
\acrodef{FPN}{Feature Pyramid Network}

\title{Digital Twin–Driven Textile Classification and Foreign Object Recognition in Automated Sorting Systems}

\author{%
  Serkan Ergun \thanks{Serkan Ergun and Tobias Mitterer contributed equally.} \\
  Institute of Smart Systems Technologies\\
  University of Klagenfurt\\
  Klagenfurt am Wörthersee, A 9020 \\
  \texttt{serkan.ergun@aau.at} \\
\And
Coauthor \\
  Tobias Mitterer
  \\
   Institute of Smart Systems Technologies\\
  University of Klagenfurt\\
  Klagenfurt am Wörthersee, A 9020 \\
  \texttt{tobias.mitterer@aau.at} \\
\And
Coauthor \\
  Hubert Zangl \\
  Institute of Smart Systems Technologies\\
  University of Klagenfurt\\
  \vspace{0.3cm}
  Klagenfurt am Wörthersee, A 9020 \\
  \vspace{0.3cm}
  and \\
  AAU SAL USE Laboratory \\
  Silicon Austria Labs \\ 
  Klagenfurt am Wörthersee, A 9020\\
  \texttt{hubert.zangl@aau.at} \\
}

\begin{document}

\maketitle

\begin{abstract}
The increasing demand for sustainable textile recycling requires robust automation solutions capable of handling deformable garments and detecting foreign objects in cluttered environments. This work presents a digital twin–driven robotic sorting system that integrates grasp prediction, multi-modal perception, and semantic reasoning for real-world textile classification. A dual-arm robotic cell equipped with RGB-D sensing, capacitive tactile feedback, and collision-aware motion planning autonomously separates garments from an unsorted basket, transfers them to an inspection zone, and classifies them using state-of-the-art \acp{VLM}.

We benchmark nine \acp{VLM} from five model families on a dataset of 223 inspection scenarios comprising shirts, socks, trousers, underwear, foreign objects (including garments outside of the aforementioned classes), and empty scenes. The evaluation assesses per-class accuracy, hallucination behavior, and computational performance under practical hardware constraints. Results show that the Qwen model family achieves the highest overall accuracy (up to \SI{87.9}{\percent}), with strong foreign object detection performance, while lighter models such as Gemma3 offer competitive speed–accuracy trade-offs for edge deployment.

A digital twin combined with MoveIt enables collision-aware path planning and integrates segmented 3D point clouds of inspected garments into the virtual environment for improved manipulation reliability. The presented system demonstrates the feasibility of combining semantic \ac{VLM} reasoning with conventional grasp detection and digital twin technology for scalable, autonomous textile sorting in realistic industrial settings.
\end{abstract}

\section{Introduction}
Robotic manipulation of deformable objects remains one of the central open challenges in automation \cite{Herguedas2019_deformable_survey, Billard2019_robot_manipulation_trends}. Among deformable materials, garments represent a particularly demanding subset due to high intra-class variability, self-occlusion, non-rigid dynamics, and frequent entanglement in bulk scenarios. These challenges are amplified in textile recycling scenarios, where garments are typically presented as unordered heaps and may contain foreign objects such as plastic packaging, metallic accessories, or non-textile waste. Reliable perception and manipulation under such conditions require tight integration of sensing, reasoning, motion planning, and safety mechanisms. From a regulatory perspective, the European Union’s upcoming Digital Product Passport (DPP) for textiles, expected to become mandatory by the end of 2027, aims to enhance traceability and material transparency \cite{eu_dpp_textiles, eu_recycling}. However, legacy textiles and non-registered garments will remain in recycling streams for years, necessitating perception-driven classification and inspection capabilities.

Recent progress in multi-modal neural networks has introduced \acp{VLM}, which combines vision and language tasks to enable semantic queries on images \cite{radford2021_CLIP}. In \cite{Ergun2025_textile_vlm}, the authors demonstrate the feasibility of integrating \acp{VLM} with \acp{CNN} for garment classification and sorting tasks. The sorting system was running in a lab setup and did not address scalability, robustness to larger garments. and foreign objects being present in the garment heap. 
This work extends the approach presented in \cite{Ergun2025_textile_vlm} toward a robust, multi-stage robotic sorting architecture suitable for realistic industrial scenarios. In this work, we present a benchmark evaluating the performance of different \acp{VLM} on textile classification and foreign object detection tasks. The robotic setup is extended to a dual-robot cell comprising two UR7e manipulators, enabling the handling of larger garments and facilitating precise, controlled, and safe manipulation. Furthermore, the system incorporates a digital twin of the grasping environment for collision-aware path planning. A 3D reconstruction of the currently inspected textile is integrated into the virtual scene, allowing improved grasp planning and more reliable execution.

\section{Related Work}

Recent progress in textile grasping and classification highlights a transition from pre-trained networks e.g. \acp{CNN}, relying on a predefined set of classes for training and detection, towards more inclusive, semantic-oriented methods such as \acp{VLM} or in a next step \acp{VLA}. Pre-trained networks like \cite{suchi2019easylabel} demonstrate the strong performance of \acp{CNN} with minimal hardware on their pre-trained datasets. \acp{VLM} in contrast work with techniques like Zero-shot object detection, using semantics and free-text inputs on multi-modal models to convert language inputs into the detection of specific objects in a scene. Some examples of early \ac{VLM} models are \ac{OwlVit} \cite{mindererowlvit}, applying image-text models to open-vocabulary object detection, and DINO \cite{zhang2023dino}, incorporating enhanced de-noising strategies and anchor boxes.
Additionally, there are\acp{VLM} with bigger datasets and a larger language model part, e.g. \cite{llama32} and the newer \cite{llama4} by Meta Inc or \cite{Qwen3VL} by Alibaba Cloud. These models are computation-heavy, with bigger versions requiring VRAM greater than 100 GiB but also supporting better visual reasoning capabilities. Another lightweight model with a bigger focus on the language and semantic part is Gemma3 \cite{gemma3} by Google LLC. Other examples of lightweight model are \cite{llava}, building on a \ac{LLM} and feeding it with a vision encoder and \cite{yao2024minicpm}, classified as \ac{MLLM} and able to work with multi-image input. 
Another approach by Meta AI is \ac{SAM} version 3, which takes classes and an image as input and segments the image based on detected objects, which works better if searching for specific objects, e.g. "sock", instead of more abstract concepts, e.g. "clothing". 
In robotics, once an object has been detected using sensors such as cameras, the next step is to identify suitable grasping positions. Pre-trained methods, like Ainetter et al. \cite{ainetter2021end}, already combine grasp detection with detailed pixel-wise semantic segmentation, with an example given in \cite{Ergun2023Grasping}. The above-mentioned \acp{VLM} do not contain any segmentation or grasp position detection. \ac{SAM} is directly doing segmentation and could be used as a first step in finding grasp positions. Finding correct grasp positions is of high importance for objects with unique shapes or materials, e.g. clothing. In this respect, \acp{CNN} have better performance as they are specifically trained for this task, but there are \acp{VLM} developed, which take semantic input, e.g. "the handle of the cup", and return grasp positions, \cite{Huang2024VLMandLLMSemantic, mirjalili2024langrasp}. Current approaches are combining the vision, grasp detection and robot control parts into one model. These are labeled as \ac{VLA}, where a semantic input in combination with vision sensors are directly mapped to robot actions, e.g. "grab the cup on the table". \ac{VLA} approaches include \cite{kawaharazuka2025_VLA_review}, which gives a review on how \acp{VLA} work and how they developed. OpenVLA \cite{kim24openvla} is based on Llama 2, and trained for a given set of available robots, with the option to adapt to new robots via parameter-efficient fine-tuning. Another \ac{VLA} example is Gr00t n1 \cite{nvidia2025gr00tn1openfoundation} by NVIDIA, which uses a dual-system model where a \ac{VLM} is responsible for vision and semantic detection and a diffusion transformer model is responsible to transform it into commands for the robot. Gr00t n1 has a direct connection to NVIDIA's simulation environment "Isaac" which contains models for most commercial robots.
 
Another important part in robot grasping scenarios is knowing the environment of the robot and being able to do movement planning of the robot with active collision avoidance. One approach to do that is using a Digital Twin of the robot for path and movement planning before giving the actual movement command to the real robot. An overview of using Digital Twins in robot grasping is given in \cite{Mazumder2023}. In this context, another important aspect is knowing the object to be grasped and, in turn, having a point-cloud representation of the object inside the digital twin. \cite{Schoenberger2016_object_reconstruction} is an early work of creating a 3D presentation of an object from multiple \ac{RGB} images.

 Our method uniquely applies \acp{VLM} alongside a \ac{CNN} within an environment for sorting textiles from a dense pile in a box based on defined semantic classes of type and does a comprehensive study on the performance of available \acp{VLM} in regards with detecting correct classes and detecting foreign objects in the pile of garments. Additionally, our approach applies a digital twin strategy with MoveIt for path planning of the robots, where a 3D reconstruction of the inspected textile is added to the digital twin.
\section{Experimental Setup}
\begin{figure} [ht!]
    \centering
    \includegraphics[width=1.0\linewidth]{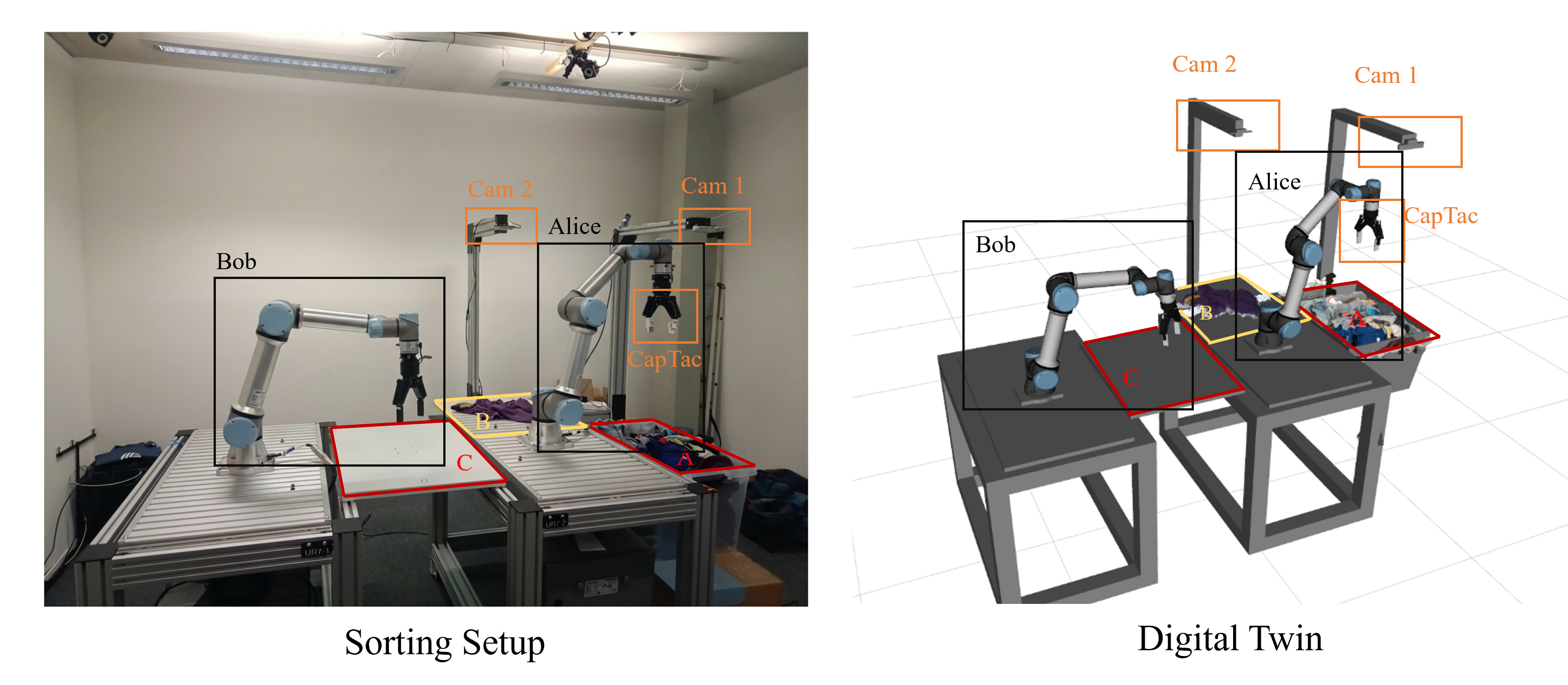}
    \caption{An overview of the experimental setup and the corresponding digitial twin in RViz.}
    \label{fig:experimental setup}
\end{figure}

\begin{figure} [h]
    \centering
    \includegraphics[width=1.0\linewidth]{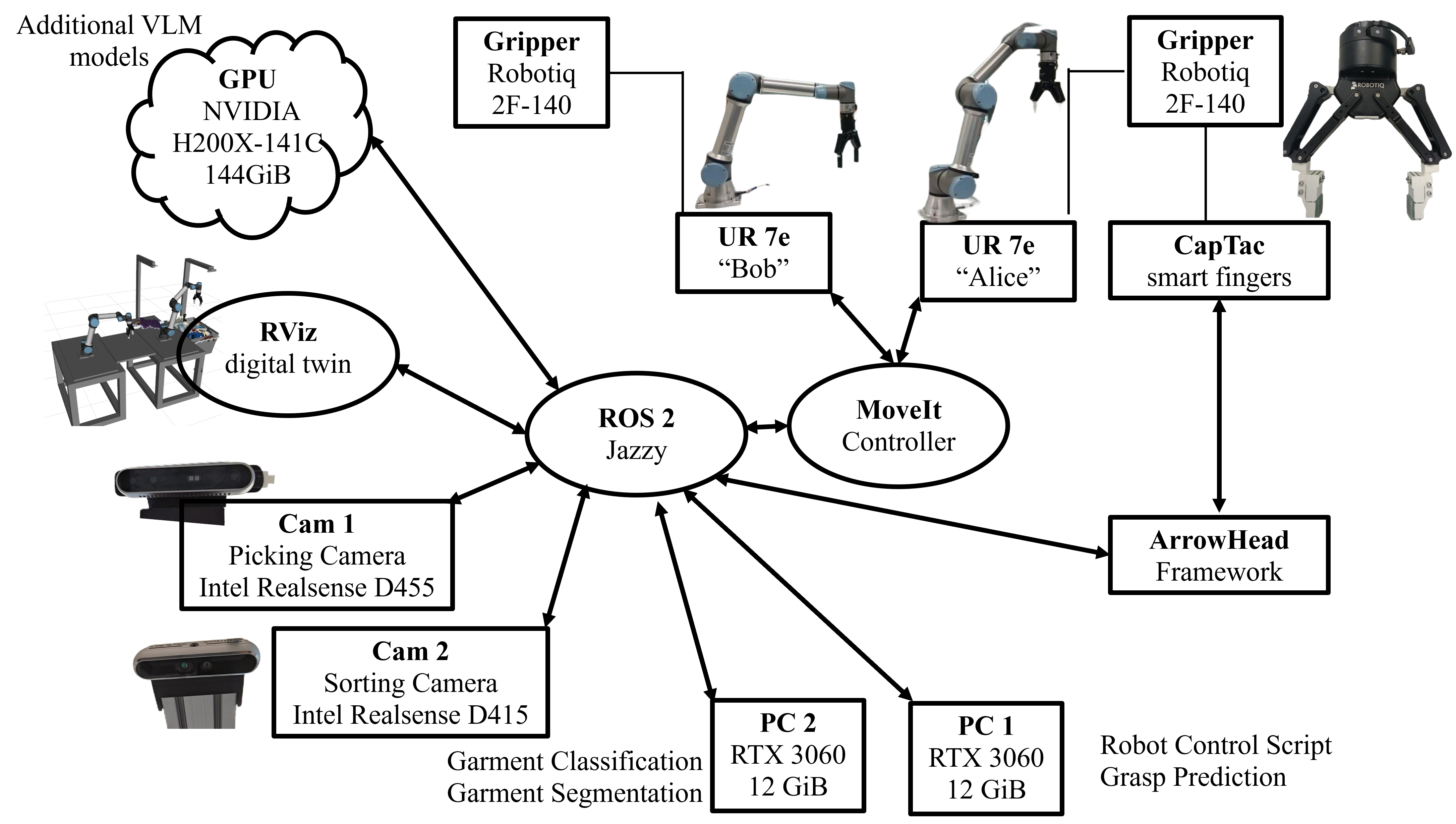}
    \caption{Schematic of our framework, consisting of two UR7e robots (\textit{Alice} and \textit{Bob} with Robotiq 2F-140 grippers, \textit{Alice} uses CapTac as fingertips.), two home grade desktop PC's equipped with Nvidia RTX 3060 budget consumer graphics cards, one Nvidia H200X professional graphics card and two Intel Realsense cameras for grasp detection and object classification. }
    \label{fig:framework}
\end{figure}

A side-by-side view of the experimental setup and its digital twin in RViz is shown in Figure \ref{fig:experimental setup}. The interconnection of all individual components is shown in Figure \ref{fig:framework}.
Our garment manipulation and sorting setup consists of two identical UR7e robots (\textit{Alice} and \textit{Bob}) placed at a distance of \SI{1.4}{\meter}. Using a modified version of the grasp prediction algorithm by \cite{ainetter2021end}, which takes depth and RGB images of \textit{Cam 1} that runs on a lab PC (\textit{PC 1}) with an 11th Gen Intel® Core™ i7-11700KF @ 3.60GHz × 16 CPU with 64 GB of DDR4 RAM paired with a \ac{GPU}  NVidia GeForce RTX 3060 with 12GiB V-RAM, \textit{Alice} picks up an item and places from zone \textit{A} and places it in zone \textit{B}  for inspection. A visual language model, which runs on a separate, but identical PC (\textit{PC 2}) using the RGB image stream of \textit{Cam 2}, classifies the object either as trousers, sock, shirt, underwear, other (other garment classes or foreign objects) or returns empty, if it finds that zone \textit{B} is empty. Alternatively, a Nvidia H200X-141C cloud GPU with 144GiB V-RAM is used to run other models for benchmarking in the background. The same grasp prediction algorithm is then called once again to find a suitable grasping pose for robot \textit{Alice}. \textit{Alice} then picks up the item and places it in zone \textit{C}, either in dedicated containers (not shown) or within the work envelope of robot \textit{Bob} for further processing.

ROS 2 Jazzy handles the data transmission between all components and the digital twin.
The interconnection of all individual components is shown in Figure \ref{fig:framework}. The control script overseeing the entire procedure runs on \textit{PC 1}. The robot trajectories including obstacle avoidance (objects taken from the digital twin) are planned by MoveIt. The depth and RGB stream of \textit{Cam 1} can be streamed into the digital twin (see Figure \ref{fig:experimental setup} -left). Objects in zone \textit{B} are, in parallel, segmented as pointclouds. These pointclouds can be stored and converted to *.stl files (with required post-processing), which could be used for further simulations or model training. The segmented pointcloud is also displayed on the digital twin (see Figure \ref{fig:experimental setup} -left).

The capacitive finger sensors are connected to the ROS2 system via the Arrowhead Eclipse framework \cite{Varga2017_arrowhead_eclipse}, in which the fingers are defined as a capacitive sensor, which provides normal and shear forces measured on multiple channels to the system. On the ROS2 side, an Arrowhead Consumer is running, which takes the measured forces and publishes them into the ROS2 system. The Arrowhead system is used to be able to transmit the sensor values in a safe and secure transmission channel, with metadata describing the sensor itself to allow for the sensor front-end to be exchanged if needed without the need to change the system backend.

\section{Experimental Procedure}
\begin{figure} [ht!]
    \centering
    \includegraphics[width=0.8\linewidth]{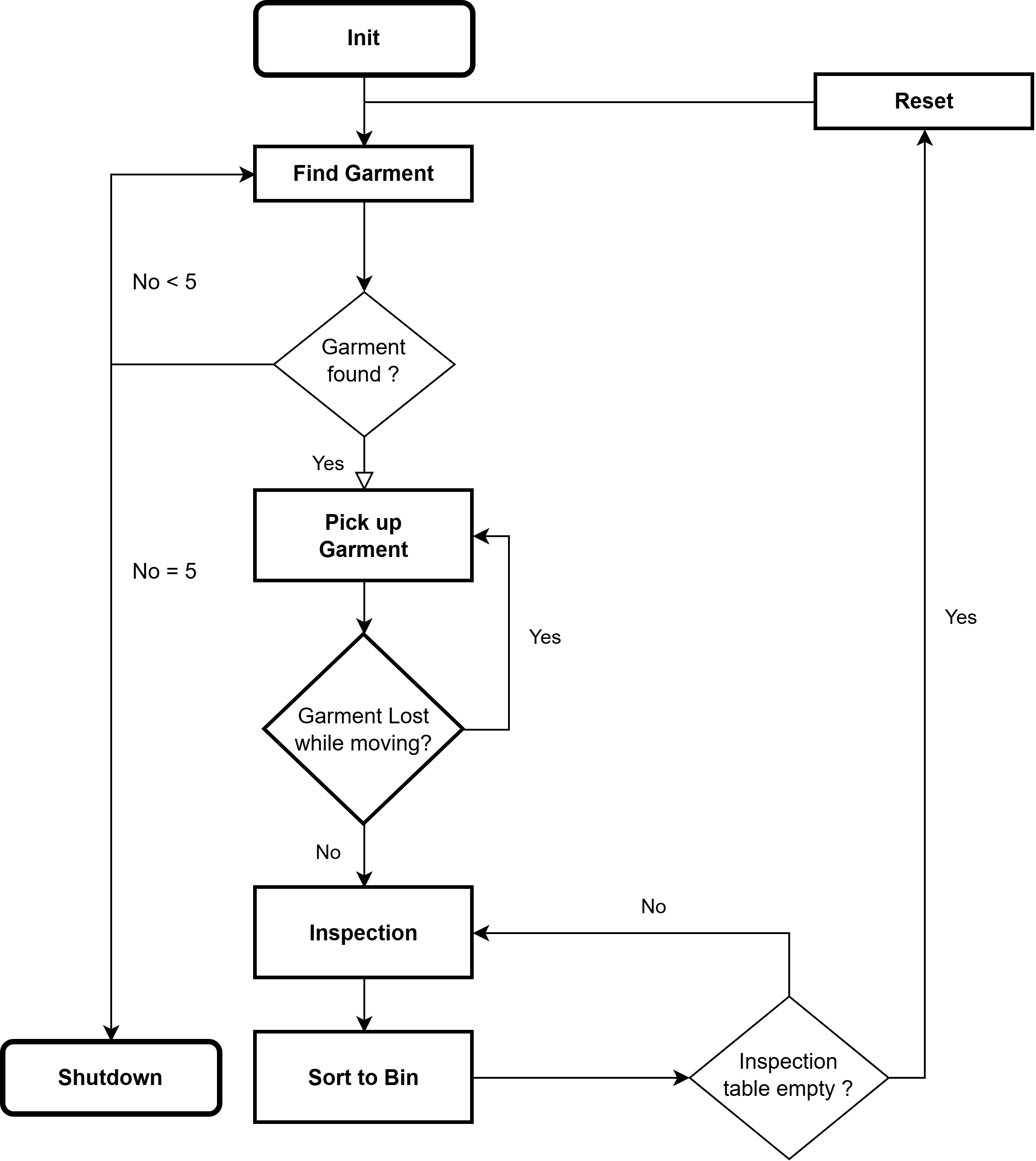}
    \caption{Flowchart of the textile inspection process.}
    \label{fig:inspection_flowchart}
\end{figure}

This section illustrates the experimental sorting procedure in detail. Before the automated garment sorting is started, a subset of random clean items is thrown randomly in a transparent basket alongside random foreign objects, such as bottles, cans and objects from the EGAD training set \cite{EGAD} and placed in the \ac{FOV} of \textit{Cam 1}. The basket does not need to be aligned, as a dynamic bounding box for the region of interest (objects inside the basket) is set, to avoid detection of items outside of the basket. 
Next, the control script is started. A simplified flowchart describing the procedure is shown in Figure \ref{fig:inspection_flowchart}. Robot \textit{Alice} initializes by moving to a safe position outside the \acp{FOV} of the cameras and the baseline of the CapTac sensors is recorded. Then a suitable grasping candidate is selected by the grasp prediction algorithm (the control script sends a ROS2 service call to the grasp prediction script). If no immediate candidate is found, up to four additional request are sent. If all attempts return no candidates, it is assumed, that no more garments are inside the basket and the robot is shut down.
If a candidate in zone \textit{A} is detected, the relative pose of the target is transformed to the world coordinate frame and the inverse kinematics to reach it, is calculated and the reachability is verified by MoveIt. If it is not reachable, another grasp target is requested. Similarly, all further robot movements are also checked by MoveIt.
\textit{Alice} then moves to the target and picks up the garment The grasp success is then determined by the feedback of the CapTac sensors \cite{CapTac}. Spacers atop the sensors avoid full closure of the gripper, in case no item is grasped. \texttt{Alice} then lifts the garment and performs a shaking motion with its wrist joints to loosen up the garment and drop possible by-catch above the basket. The garment is pulled over the edge of the inspection table to induce passive spreading, allowing it to lie as flat and extended as possible on the table surface.
The object is then placed in the inspection zone \textit{B} using the same motion planner. \textit{Alice} will move outside the inspection zone and the garment classification script, which utilizes a locally run \ac{VLM} using the Python API of Ollama \cite{ollama}, is called by another ROS2 service call. As mentioned above, the considered garment classes are: trousers, shirt, underwear, sock. Any foreign object and unidentifiable garment is classified as: other. If no object is detected, the script should return: empty. If an object has been classified, a grasp position for the given object is again found with the grasp prediction algorithm and the object is moved to zone \textit{C} for further processing, which could be either further inspection or handover to Bob, which sorts it into the respective bins behind Bob on the ground. Next, the grasp prediction algorithm, as it is significantly faster (about 150 ms), than running the classification \ac{VLM}, checks zone \textit{B} if there is still an object to be classified and sorted. If no object to be sorted is found, the system resets and tries to find a new textile in zone \textit{A}.
A minimal Python script showcasing the usage of Ollama with Python3 is shown in listing \ref{list:ollama_code}.
\begin{lstlisting}[language=Python, caption={Minimal Python code example for running a Vision Language Model with Ollama}, label=list:ollama_code]

import ollama

response = ollama.chat(
    model='model-name',
    messages=[{"role": "system",
        "content": "You are an intelligent robotic arm."},
        {'role': 'user', 'content': '"Do you spot a clothing item on the table? "
    "If yes: Classify them in the classes: "
    "shirt, sock, underwear or trousers. "
    "Do you see something else instead? respond with other. "
    "Is the table empty? respond with empty. "
    "Your response is a single word - either "
    "shirt, sock, underwear, trousers, other or empty"', 'images': [fullPathToImages]
            }]
)
\end{lstlisting}

The segmentation script takes RGB and depth images of \textit{Cam 2} as input. During initialization, set of images of the empty inspection table (zone \textit{B}) are taken and used as a baseline for later incoming ROS2 service calls. Fresh RGB and depth images are then subtracted from the baseline. Fixed thresholds (depth: >\SI{5}{\milli\metre}, RGB: >15 change in any of the three color streams) are used to return a pointcloud with RGB information. This basic approach is sufficient for displaying the segmented object in RViz, but requires post-processing (smoothing and interpolation of surfaces) to export proper 3D models, which is outside of the scope of the present work.
\section{Results}
In total, 219 items (garments and foreign objects) were grasped from the basket and placed on the inspection table for classification. To address robustness, we chose to run the \acp{VLM} using uncropped images, such that parts of the floor with distracting items are present.

\begin{figure}[h!]
    \centering
    \begin{subfigure}{0.48\textwidth}
        \centering
        \includegraphics[width=\linewidth]{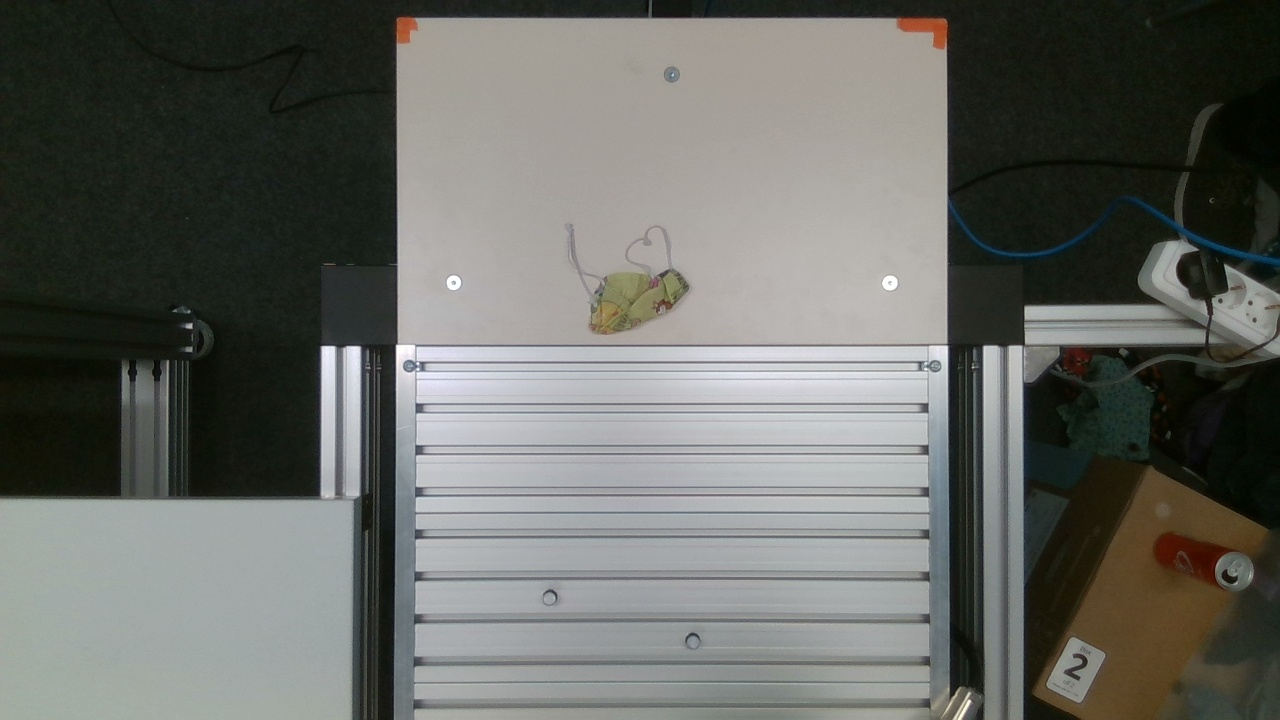}
        \caption{Class: "Other", gemma3:12b: "Other", llama3.2-vision:90b: "Other", llama4:16x17b: "Other", llava:34b: "Other", minicpm-v:8b: "Sock", ,qwen3-vl:235b: "Other", qwen3-vl:8b: "Other", qwen3.5:35b: "Other", qwen3.5:122b: "Other" }
        \label{fig:garment1}
    \end{subfigure}
    \hfill
    \begin{subfigure}{0.48\textwidth}
        \centering
        \includegraphics[width=\linewidth]{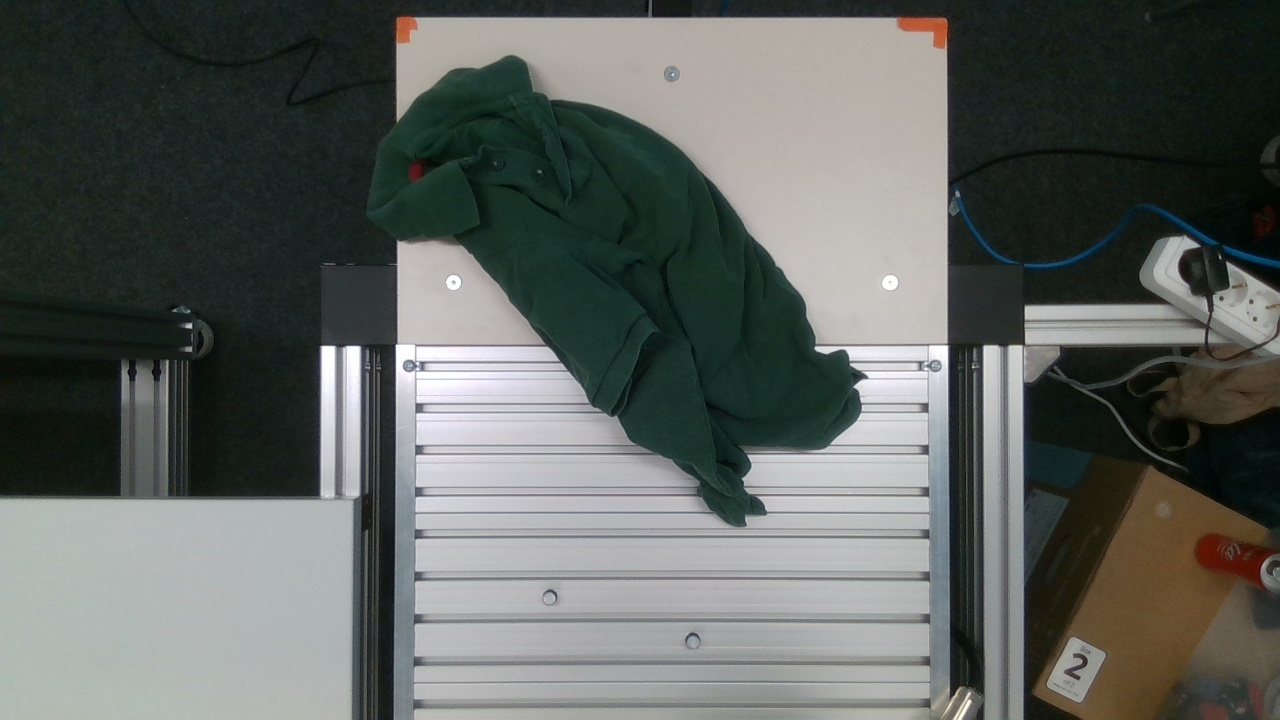}
        \caption{Class: "Shirt", gemma3:12b: "Other", llama3.2-vision:90b: "Other", llama4:16x17b: "Shirt", llava:34b: "The image shows a piece of green fabric ..... industrial machine or equipment.", minicpm-v:8b: "Shirt", ,qwen3-vl:235b: "Shirt", qwen3-vl:8b: "Shirt", qwen3.5:35b: "Shirt", qwen3.5:122b: "Shirt" }
        \label{fig:garment2}
    \end{subfigure}
    \caption{Two exemplary images of garments in zone \texttt{B}. To consider robustness in cluttered environments, additional smaller garments were placed on the ground and several distracting items were present. The response of each model is given together with its ground truth.}
    \label{fig:two_garments}
\end{figure}

The distribution to each class is shown in Table \ref{tab:model_results} below. Due to combination of a well-known and robust grasp prediction algorithm and capacitive tactile fingertip sensors to detect immediate fail-grasps and object loss, zone \textit{B} was not empty during our experiments. To investigate the behavior of \acp{VLM} towards potential hallucinations, items were removed, before the inspection algorithm was called and another two empty inspections service calls where sent after the experimental run. The comparatively light-weight model \texttt{gemma3:12b} was used in the live experiments and ran on \textit{PC 2} alongside the object segmentation for displaying the real object on the digital twin. All further models were later were later run either on of the identical lab PC's or the H200 Cloud GPU. The accuracies for each model and garment class are listed in Table \ref{tab:model_results}. The model family \texttt{gwen3.5} was just released shortly before the submission deadline and also included in this benchmark.

\begin{figure}[h]
    \centering
    
    \begin{subfigure}[b]{0.3\textwidth}
        \centering
        \includegraphics[width=\textwidth]{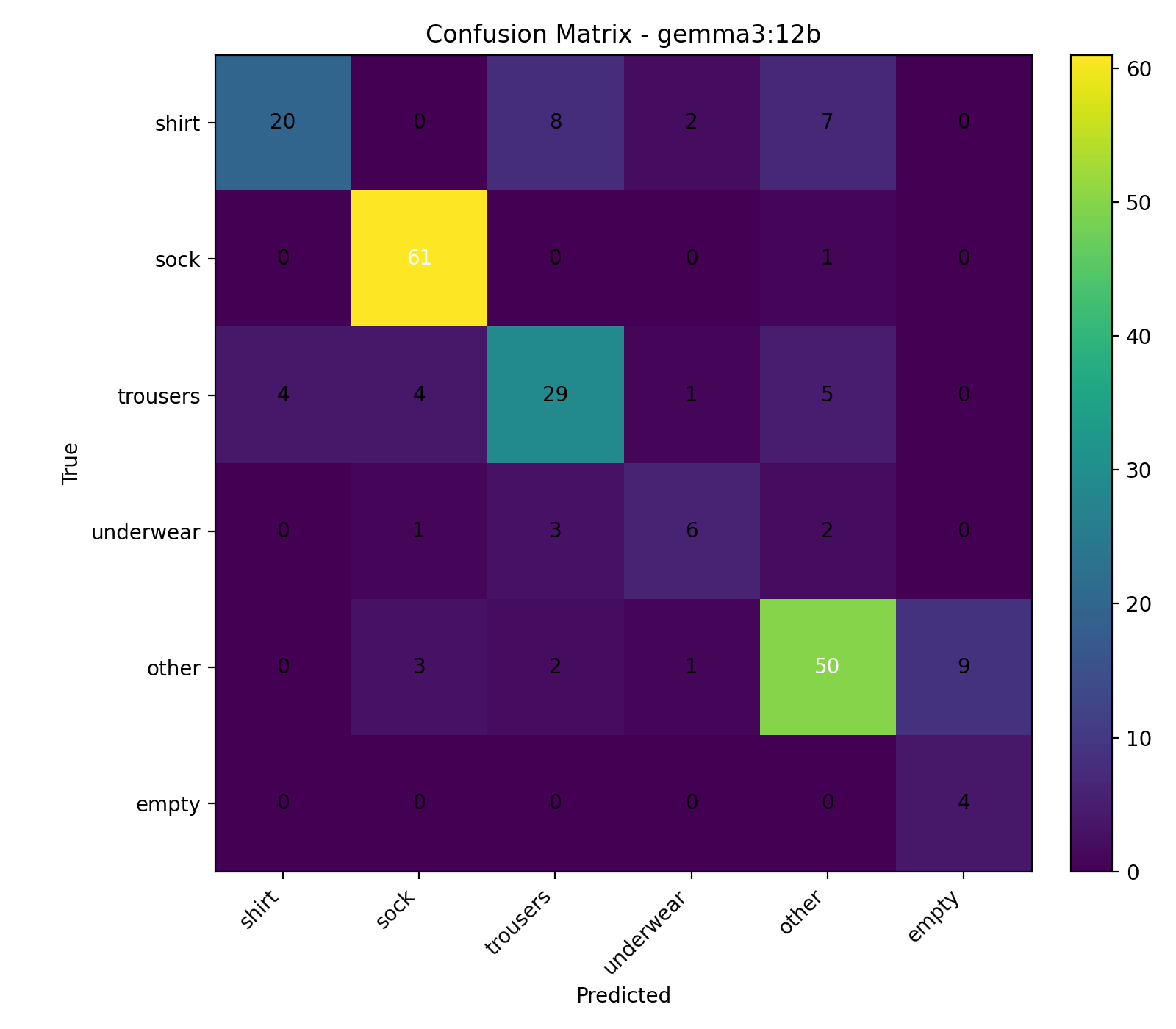}
        \caption{gemma3\_12b}
    \end{subfigure}
    \hfill
    \begin{subfigure}[b]{0.3\textwidth}
        \centering
        \includegraphics[width=\textwidth]{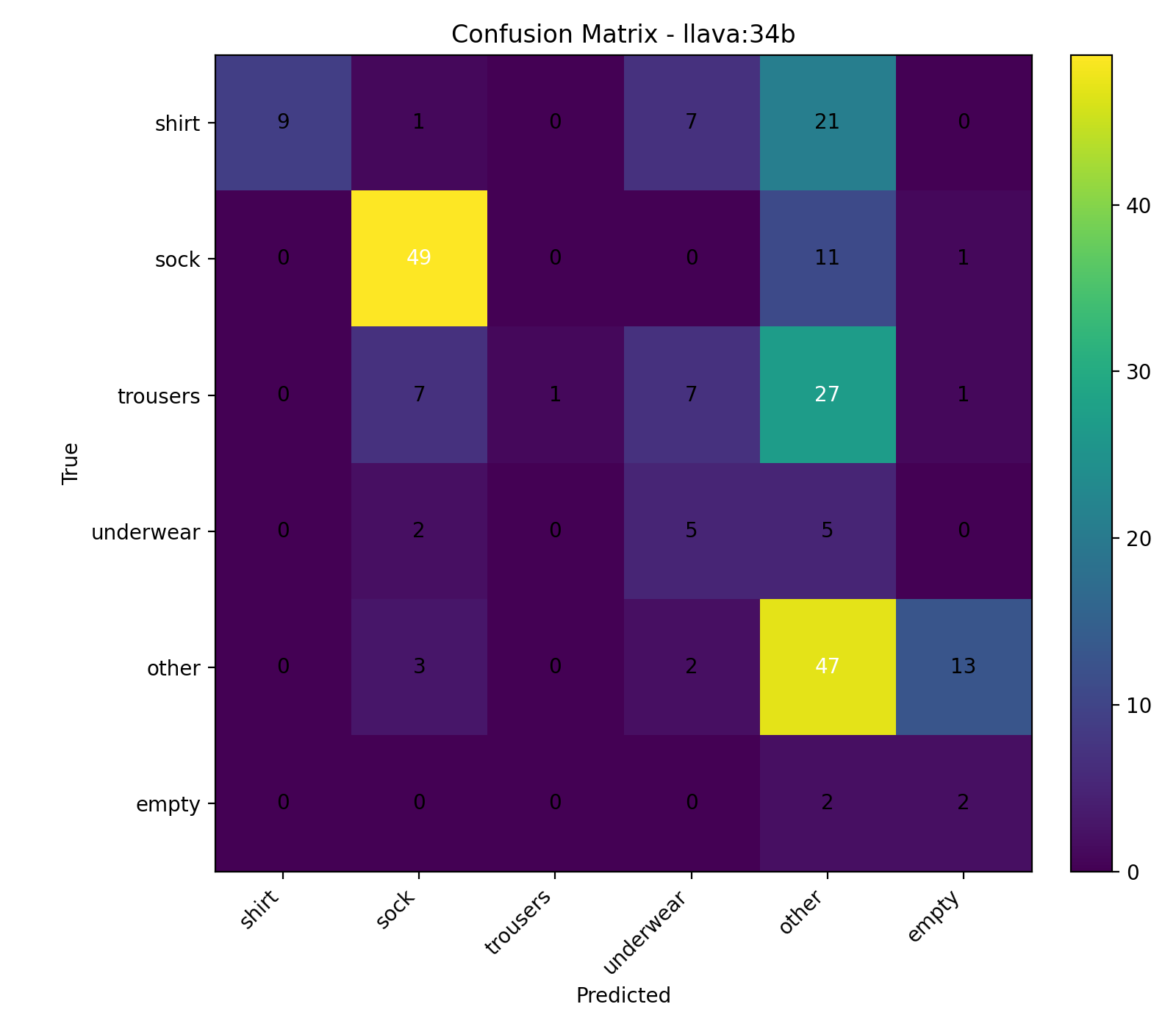}
        \caption{llava\_34b}
    \end{subfigure}
    \hfill
    \begin{subfigure}[b]{0.3\textwidth}
        \centering
        \includegraphics[width=\textwidth]{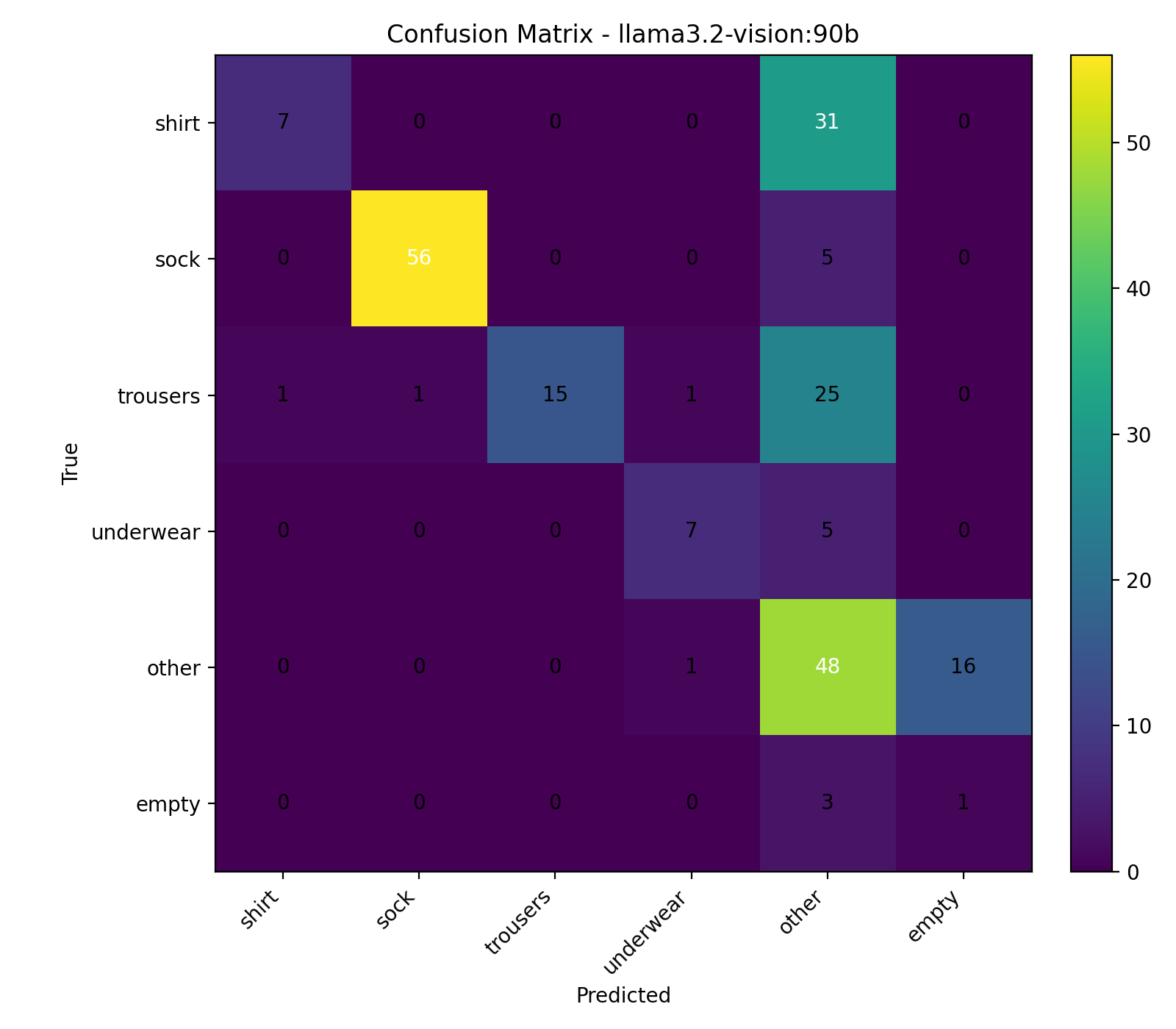}
        \caption{lama3.2-vision\_90b}
    \end{subfigure}
    
    \vspace{0.2cm}
    
    \begin{subfigure}[b]{0.3\textwidth}
        \centering
        \includegraphics[width=\textwidth]{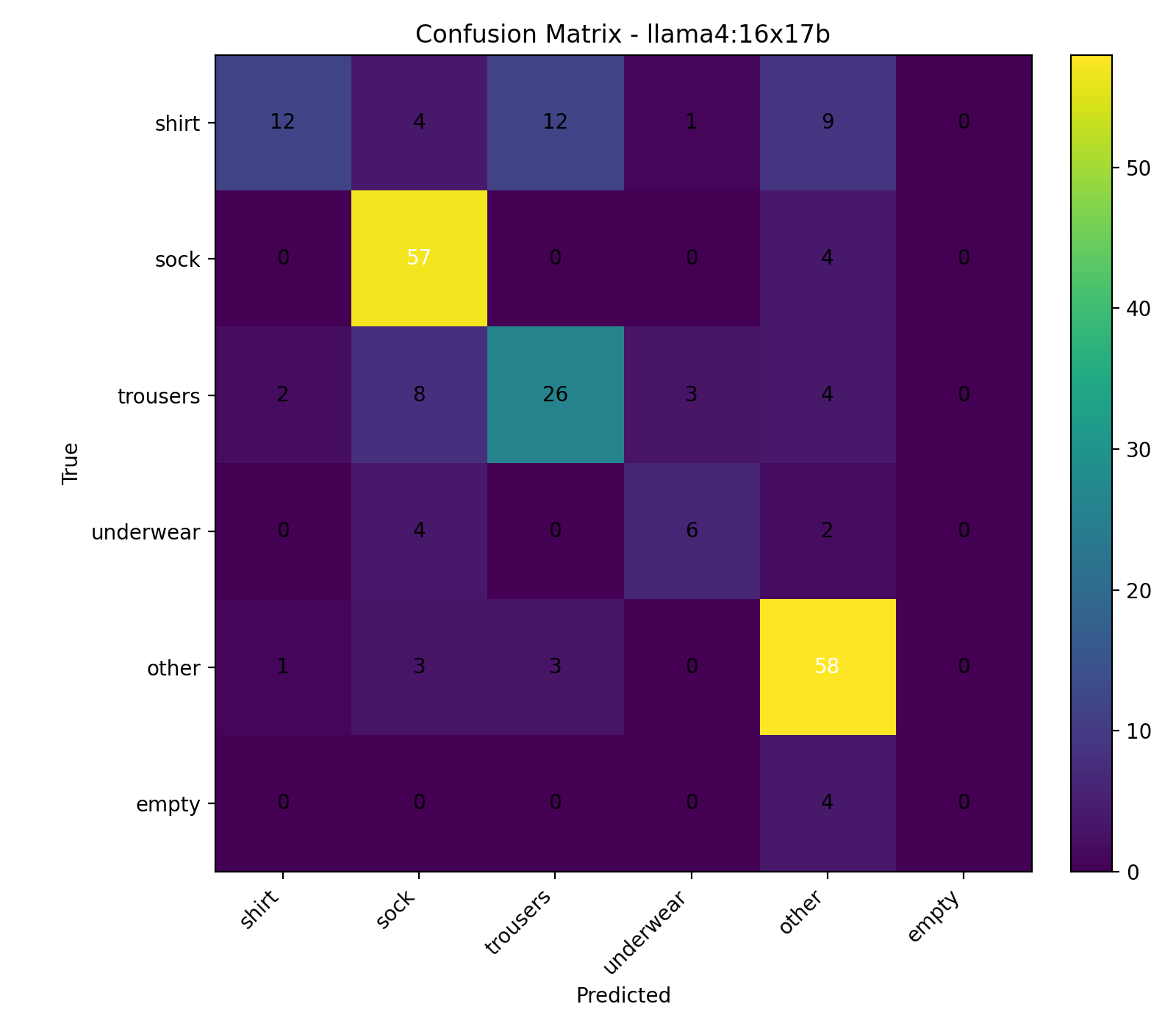}
        \caption{llama4\_16x17b}
    \end{subfigure}
    \hfill
    \begin{subfigure}[b]{0.3\textwidth}
        \centering
        \includegraphics[width=\textwidth]{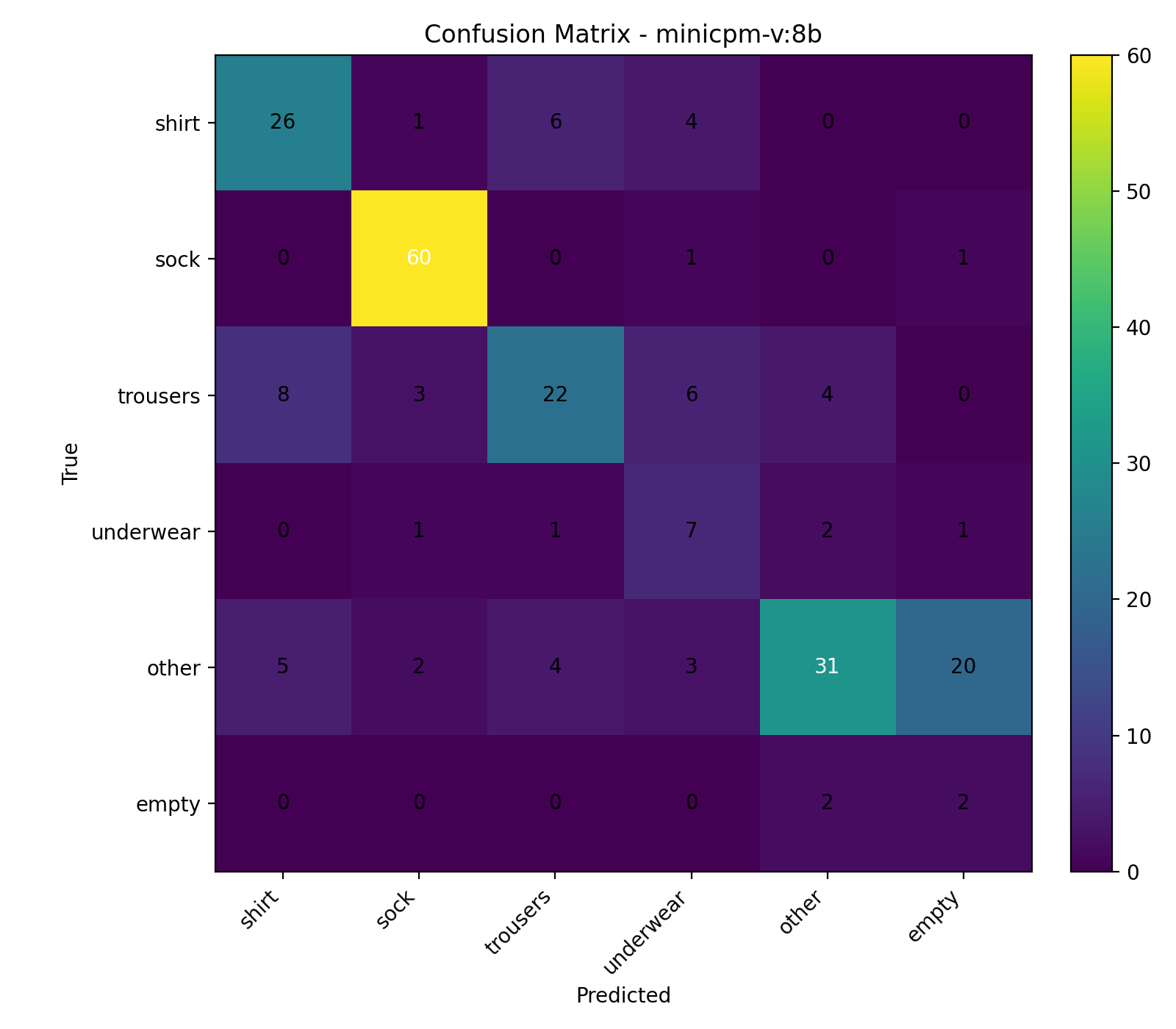}
        \caption{minicpm-v\_8b}
    \end{subfigure}
    \hfill
    \begin{subfigure}[b]{0.3\textwidth}
        \centering
        \includegraphics[width=\textwidth]{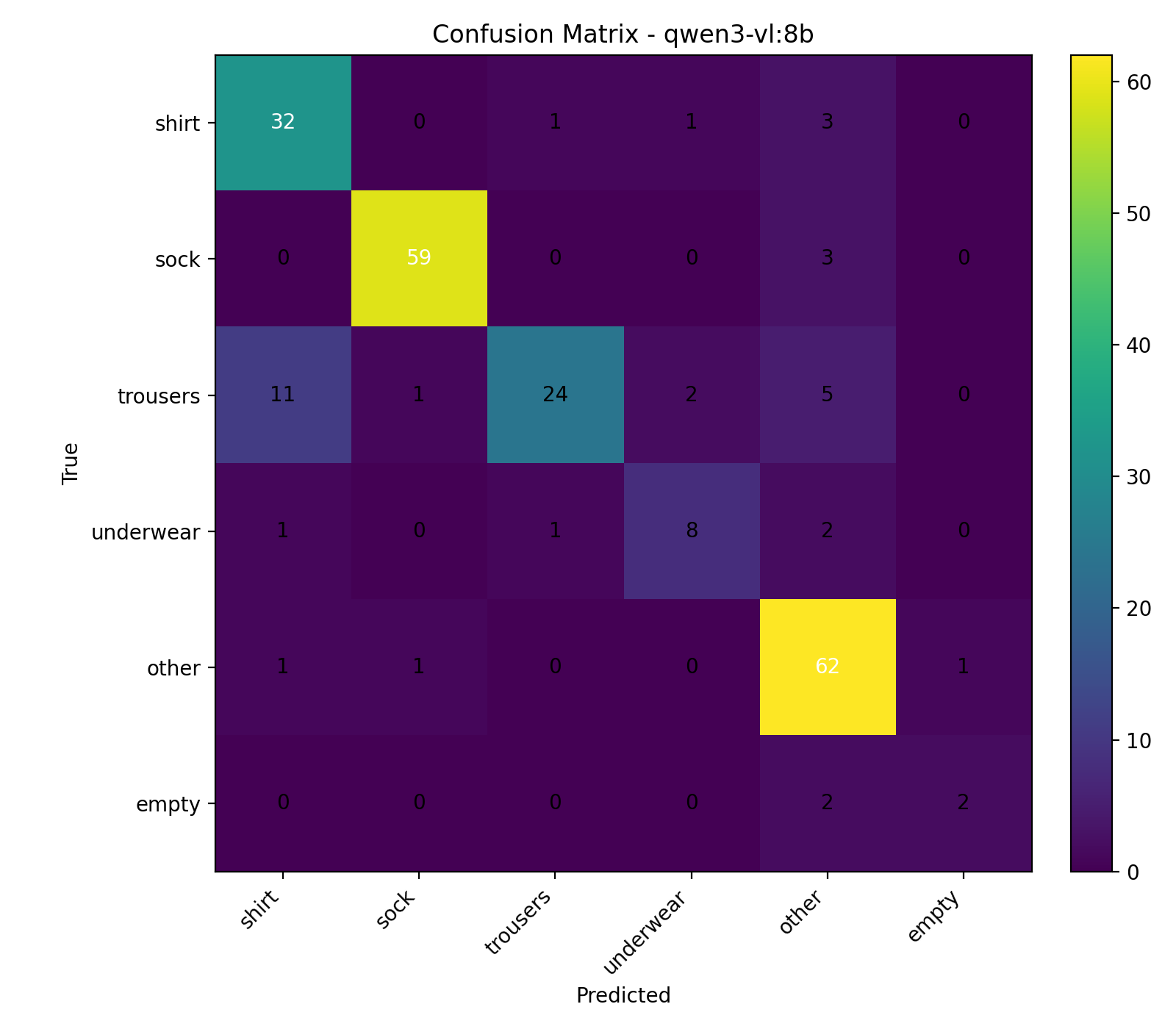}
        \caption{qwen3-vl\_8b}
    \end{subfigure}
    
    \vspace{0.2cm}
    
    \begin{subfigure}[b]{0.3\textwidth}
        \centering
        \includegraphics[width=\textwidth]{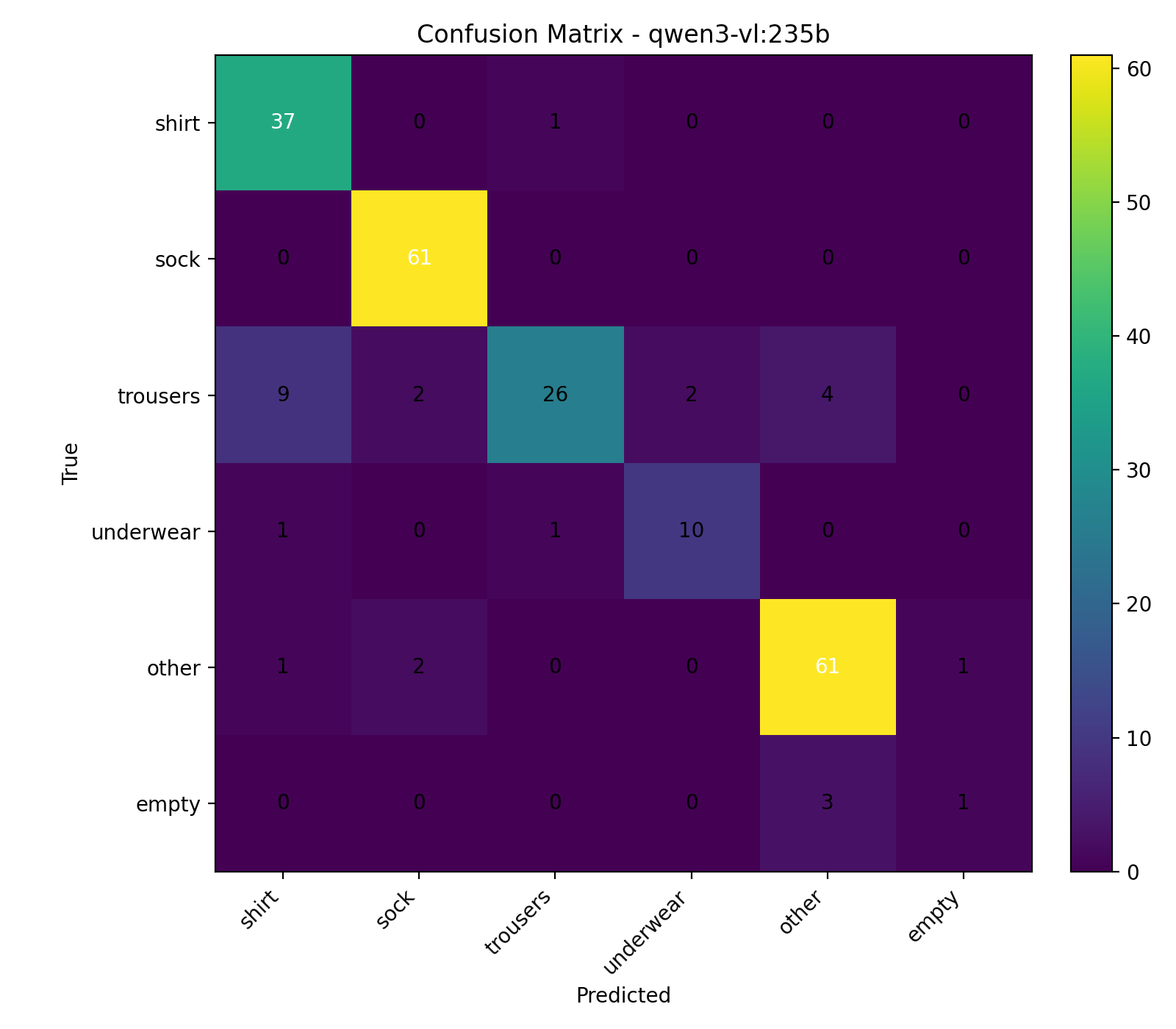}
        \caption{qwen3-vl\_235b}
    \end{subfigure}
    \hfill
    \begin{subfigure}[b]{0.3\textwidth}
        \centering
        \includegraphics[width=\textwidth]{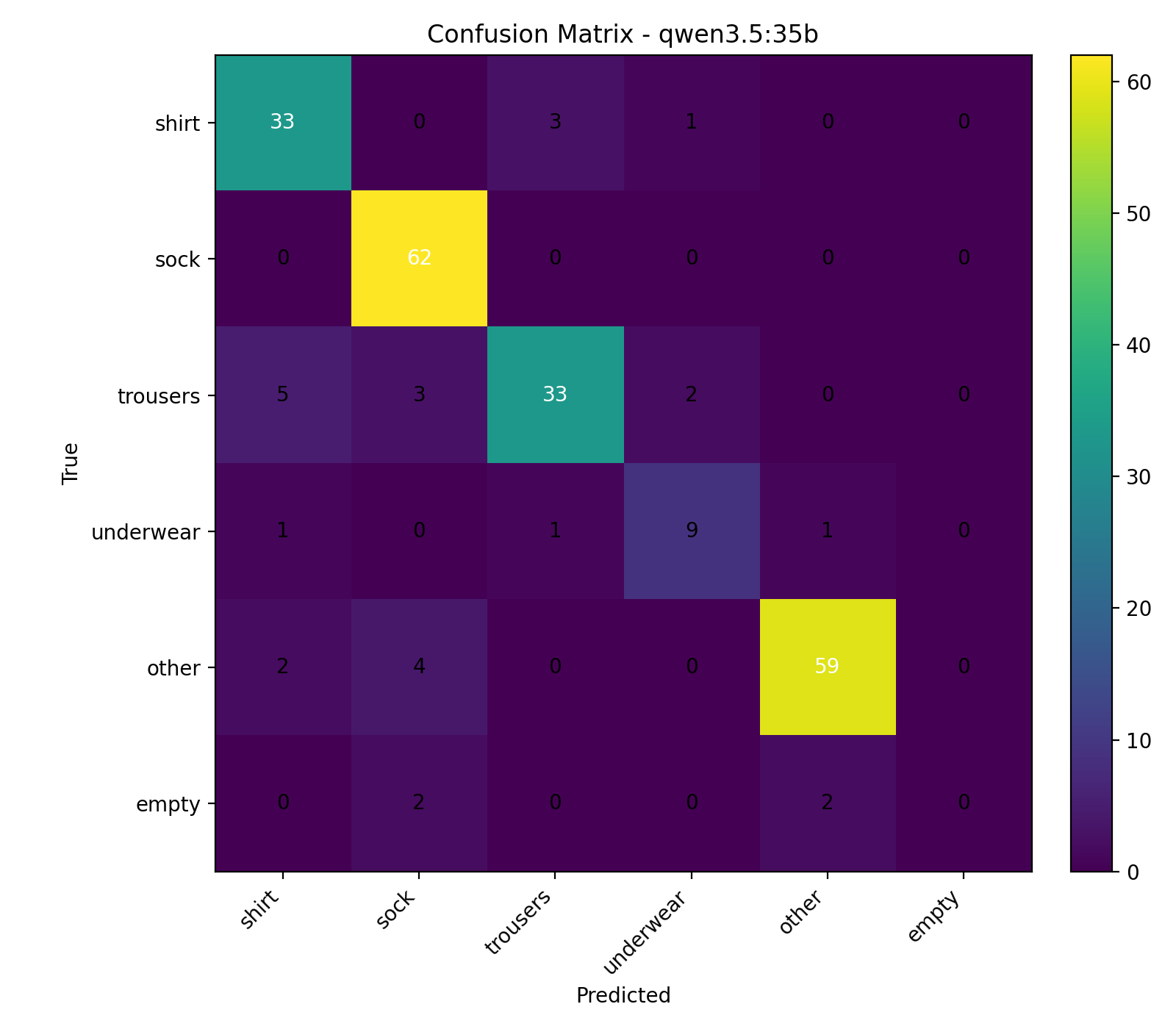}
        \caption{qwen3.5\_35b}
    \end{subfigure}
    \hfill
    \begin{subfigure}[b]{0.3\textwidth}
        \centering
        \includegraphics[width=\textwidth]{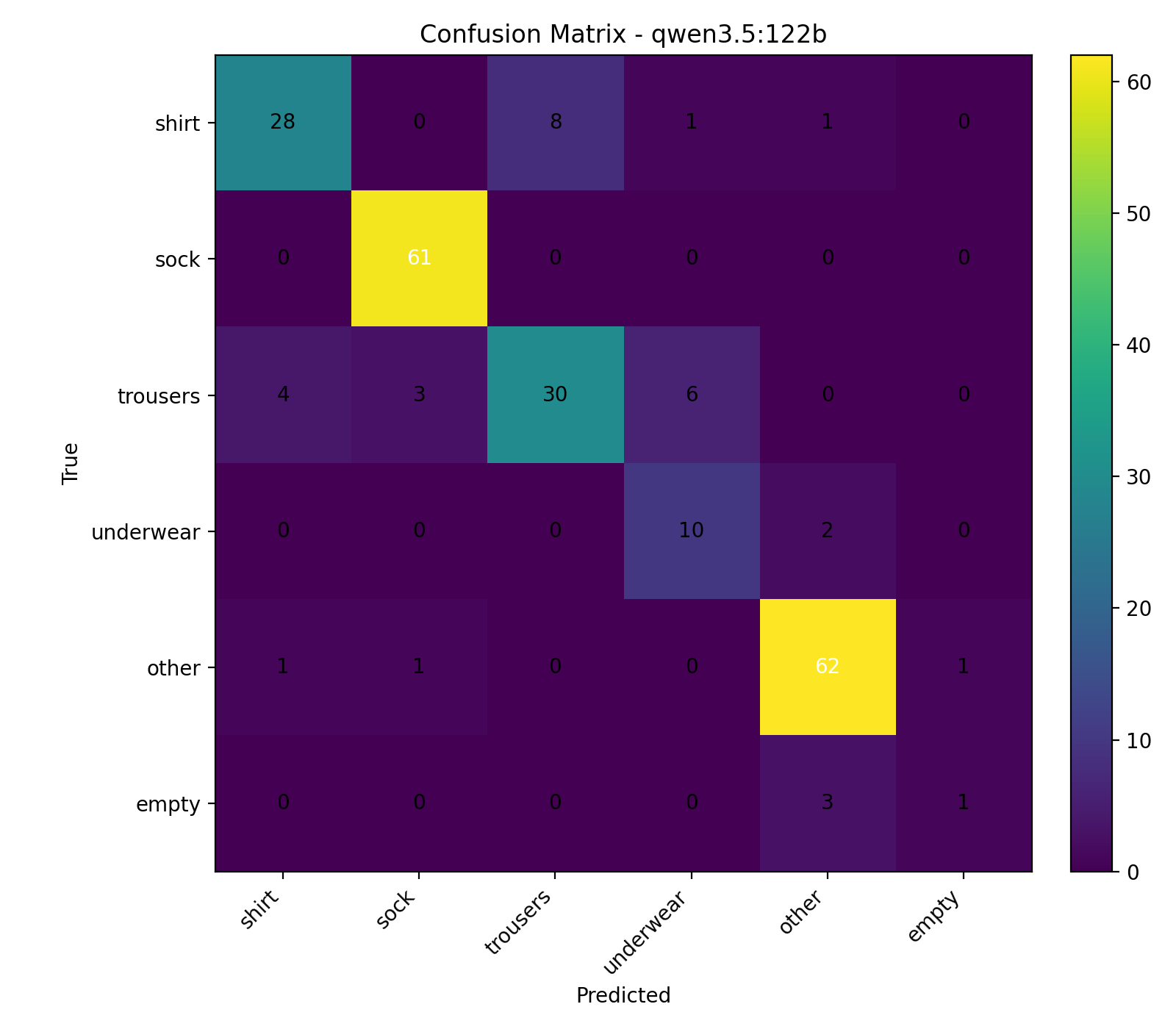}
        \caption{qwen3.5\_122b}
    \end{subfigure}
    
    \caption{Confusion matrices for all nine considered \acp{VLM}.}
    \label{fig:confusion_matrices}
\end{figure}

A prediction was considered as correct, if the response of the model was exactly matching the aforementioned classes (ignoring lower/uppercase lettering). Returning a different, but semantically correct class name was considered as wrong. Also, if the response also contained more words than prompted, the response was treated as incorrect. Especially, the model \texttt{llava\_34b} was not always able to follow this rule and often returned full sentences. Models from the \texttt{llama} family tend to start hallucinating if the garment class changes after multiple instances of the same class. It will also tends to hallucinate in sparse situations of no object being presented. 
\begin{table}[]
\centering
\caption{Performance Benchmark for all investigated models. }
\label{tab:model_results}

\begin{tabular}{lcccccccc}
\hline
            & Overall & Shirt & Sock & Trousers & Underwear & Other & Empty\\
\hline
Image Count & 223 & 38 & 64 & 43 & 12 & 65 &4  \\
\hline
Models   & \multicolumn{7}{c}{Accuracy}   \\

gemma3:12b & \SI{76.23}{\percent} & \SI{55.26}{\percent} & \SI{95.31}{\percent} &  \SI{67.44}{\percent}& \SI{50.00}{\percent} &\SI{76.92}{\percent} &\SI{100.00}{\percent} \\

llama3.2-vision:90b & \SI{60.09}{\percent} & \SI{18.42}{\percent} & \SI{87.50}{\percent} &  \SI{34.88}{\percent}& \SI{58.33}{\percent} &\SI{73.85}{\percent} &\SI{25.00}{\percent} \\

llama4:16x17b & \SI{71.30}{\percent} & \SI{31.58}{\percent} & \SI{89.06}{\percent} &  \SI{60.47}{\percent}& \SI{50.00}{\percent} &\SI{89.23}{\percent} &\SI{00.00}{\percent} \\

llava:34b & \SI{50.67}{\percent} & \SI{23.68}{\percent} & \SI{76.56}{\percent} &  \SI{2.33}{\percent}& \SI{41.67}{\percent} &\SI{72.31}{\percent} &\SI{50.00}{\percent} \\

minicpm-v:8b & \SI{65.02}{\percent} & \SI{71.05}{\percent} & \SI{95.31}{\percent} &  \SI{51.16}{\percent}& \SI{58.33}{\percent} &\SI{43.08}{\percent} &\SI{50.00}{\percent} \\

qwen3-vl:235b & \SI{87.89}{\percent} & \SI{97.37}{\percent} & \SI{100.00}{\percent} &  \SI{60.47}{\percent}& \SI{83.33}{\percent} &\SI{93.85}{\percent} &\SI{25.00}{\percent} \\

qwen3-vl:8b & \SI{83.86}{\percent} & \SI{86.84}{\percent} & \SI{93.75}{\percent} &  \SI{55.81}{\percent}& \SI{66.67}{\percent} &\SI{95.38}{\percent} &\SI{50.00}{\percent} \\

qwen3.5:35b* & \SI{87.89}{\percent} & \SI{89.47}{\percent} & \SI{100.00}{\percent} &  \SI{76.74}{\percent}& \SI{75.00}{\percent} &\SI{90.77}{\percent} &\SI{00.00}{\percent} \\

qwen3.5:122b*   & \SI{86.10}{\percent} & \SI{73.68}{\percent} & \SI{98.44}{\percent} &  \SI{69.77}{\percent}& \SI{83.33}{\percent} &\SI{95.38}{\percent} &\SI{25.00}{\percent} \\
\hline
\end{tabular}
\\
\small{Note: For qwen3.5, a non-stable devel version of Ollama had to used to run these models.}
\end{table}
In addition, the confusion matrices for all nine models are shown in Figure \ref{fig:confusion_matrices}.

Furthermore, the computation time for each model and prompt was recorded and stored alongside the model predictions. The average computation time $\overline{t}$ alongside the $10^{th}$ percentile $P_{10}$ and $90^{th}$ percentile $P_{90}$ are shown in Table \ref{tab:model_stats}. Due to the late breaking release of the \texttt{gwen3.5}, only a non stable devel version of Ollama could be used to run these models, which may be the reason for unreliable computation times. These experiments will be re-conducted upon eventual final submission.

\begin{table}[]
\centering
\caption{Computation times in \SI{}{\second} for all investigated models: average $\overline{t}$, $10^{th}$ percentile $P_{10}$ and $90^{th}$ percentile $P_{90}$  }
\label{tab:model_stats}

\begin{tabular}{lccccc}
\hline
 & gemma3:12b & llama3.2-vision:90b & llama4:16x17b & qwen3-vl:235b & llava:34b \\
\hline
GPU  & \multicolumn{5}{c}{H200}  \\
$\overline{t}$    & 0.653 & 0.953 & 0.969 & 2.444 & 0.409 \\
$P_{10}$          & 0.620& 0.620 & 0.904 &  1.739& 0.378  \\
$P_{90}$          & 0.689 & 0.690 & 1.016 &  3.072& 0.411 \\
\hline
\end{tabular}
\vspace{0.2cm}

\begin{tabular}{lccccc}
\hline
 & minicpm-v:8b & qwen3-vl:8b & qwen3.5:35b* & qwen3.5:122b*  \\
\hline
GPU              &  \multicolumn{4}{c}{H200}   \\

$\overline{t}$   & 0.461 & 1.595 & 12.299 & 20.480   \\
$P_{10}$         & 0.411 & 0.993 & 2.807 &3.892   \\
$P_{90}$         & 0.534 & 2.550& 27.266 & 55.869  \\
\hline \\

\end{tabular}
\\
\small{Note: For qwen3.5 a non-stable version of Ollama had to be used in order to run these models. This may have an effect on the computation time of these models.}
\end{table}

In general it can be noted, that larger models deliver better and more concise results, which is expected due to their larger datasets they were trained on. Computation times are not critical, in general, as long as the entire model can be loaded into the V-RAM of the GPU, and can therefore be processed at once, without relying on outside swap storage on the mainboard CPU RAM and/or the SSD/harddrive. The \texttt{qwen} model family shows a significantly better accuracy, precision and recall in comparison to other model families. The newly released \texttt{qwen3.5:35b} model delivers similar results as the much larger, but older \texttt{qwen3-vl:235b} model (24 vs 143 GB) and is therefore also executable on high end consumer graphics cards, whereas the larger model needs an high-performance computing datacenter scale GPU (such as the Nvidia H200) to run.
\section{Discussion, Conclusion and Outlook}
The present work introduces a digital twin driven approach for autonomous handling and sorting of garments with the capability of detecting foreign objects. The digital twin provides the foundation for obstacle avoidance using MoveIt. In total, nine different \acfp{VLM} from five different model families were investigated and their performance metrics are documented. Our results show that the \texttt{qwen} model consistently delivers the best results across all considered garment classes and foreign objects. Their main drawback is the comparatively higher computation time due to reasoning. \texttt{gemma3:12b} is an alternative for more rapidly and consistently processed images at a reduced accuracy. \texttt{gemma3:12b} was only the only model to correctly identify all scenarios with empty tables, however the sample count remains too low to attest high accuracy. Raw imagery of all processed objects (garments and others) are stored together with their labels for evaluating other models in the future. All data will be made publicly available after eventual publication.
As the garment manipulation upon classification was handled by a single robot (\texttt{Alice}) garments where not always placed ideally in zone \texttt{B}, effectively reducing the accurancy for garment classes of larger size. Socks, due to their smaller size and distinctive shape, gave the highest accuracies across all considered models.
The current state of our proposed setup has a big potential for further improvement. The next logical step is to demonstrate robot to robot handover tasks and spreading garments with multiple arms for in-depth inspection using camera images from multiple perspectives. In this way, the quality of the garment segmentation algorithm can be improved to get precise 3D models of garments for simulation tasks.
Additionally, the combination of multiple \acp{VLM} with weighting factors seems feasible, especially combining \texttt{qwen} models with \texttt{gemma} to increase the rate of empty table detection.

\section*{Acknowledgement}
This work has received funding from the ”Austrian Research Promotion Agency” (FFG) within the AdapTex project under grant number 899044 and by the European Commission, through the European H2020 research and innovation programme, KDT Joint Undertaking, and National Funding Authorities from 10 involved countries – including Hungary – under the research project Arrowhead fPVN with Grant Agreement no. 101111977.

{\small
\bibliographystyle{IEEEtran}
\bibliography{references}

@InProceedings{ainetter2021end,
  title={End-to-end Trainable Deep Neural Network for Robotic Grasp Detection and Semantic Segmentation from RGB},
  author={Ainetter, Stefan and Fraundorfer, Friedrich},
  booktitle={IEEE International Conference on Robotics and Automation (ICRA)},
  pages={13452--13458},
  year={2021}
}

@inproceedings{suchi2019easylabel,
  title={EasyLabel: a semi-automatic pixel-wise object annotation tool for creating robotic RGB-D datasets},
  author={Suchi, Markus and Patten, Timothy and Fischinger, David and Vincze, Markus},
  booktitle={2019 International Conference on Robotics and Automation (ICRA)},
  pages={6678--6684},
  year={2019},
  organization={IEEE}
}

@inproceedings{mindererowlvit,
author = "{Minderer, Matthias and Gritsenko, Alexey and Stone, Austin and Neumann, Maxim and Weissenborn, Dirk and Dosovitskiy, Alexey and Mahendran, Aravindh and Arnab, Anurag and Dehghani, Mostafa and Shen, Zhuoran and Wang, Xiao and Zhai, Xiaohua and Kipf, Thomas and Houlsby, Neil}",
title = "{Simple Open-Vocabulary Object Detection}",
year = {2022},
isbn = "{978-3-031-20079-3}",
publisher = "{Springer-Verlag}",
address = {Berlin, Heidelberg},
url = "{https://doi.org/10.1007/978-3-031-20080-9\_42}",
doi = "{10.1007/978-3-031-20080-9_42}",
booktitle = {Computer Vision – ECCV 2022: 17th European Conference, Tel Aviv, Israel, October 23–27, 2022, Proceedings, Part X},
pages = {728–755},
numpages = {28}
}

@inproceedings{
zhang2023dino,
title={{DINO}: {DETR} with Improved DeNoising Anchor Boxes for End-to-End Object Detection},
author={Hao Zhang and Feng Li and Shilong Liu and Lei Zhang and Hang Su and Jun Zhu and Lionel Ni and Heung-Yeung Shum},
booktitle={The Eleventh International Conference on Learning Representations },
year={2023},
url={https://openreview.net/forum?id=3mRwyG5one}
}

@INPROCEEDINGS{Ergun2023Grasping,
  author={Ergun, Serkan and Mitterer, Tobias and Khan, Sherjeel and Anandan, Narendiran and Mishra, Rishabh B. and Kosel, Jürgen and Zangl, Hubert},
  booktitle={2023 IEEE/RSJ International Conference on Intelligent Robots and Systems (IROS)}, 
  title={Wireless Capacitive Tactile Sensor Arrays for Sensitive/Delicate Robot Grasping}, 
  year={2023},
  volume={},
  number={},
  pages={10777-10784},
  doi={10.1109/IROS55552.2023.10342163}
}

@inproceedings{
mirjalili2024langrasp,
title={{LAN}-grasp: An Effective Approach to Semantic Object Grasping Using Large Language Models},
author={Reihaneh Mirjalili and Michael Krawez and Simone Silenzi and Yannik Blei and Wolfram Burgard},
booktitle={First Workshop on Vision-Language Models for Navigation and Manipulation at ICRA 2024},
year={2024},
url={https://openreview.net/forum?id=SfHjWbfW02}
}

@INPROCEEDINGS{Huang2024VLMandLLMSemantic,
  author={Huang, Jiayang and Limberg, Christian and Arshad, Syed Muhammad Nashit and Zhang, Qifeng and Li, Qiang},
  booktitle={2024 WRC Symposium on Advanced Robotics and Automation (WRC SARA)}, 
  title={Combining VLM and LLM for Enhanced Semantic Object Perception in Robotic Handover Tasks}, 
  year={2024},
  volume={},
  number={},
  pages={135-140},
  doi={10.1109/WRCSARA64167.2024.10685688}}

@misc{eu_recycling,
  author={{Directorate-General for Environment}},
  title="{COMMUNICATION FROM THE COMMISSION TO THE EUROPEAN
PARLIAMENT, THE COUNCIL, THE EUROPEAN ECONOMIC AND SOCIAL
COMMITTEE AND THE COMMITTEE OF THE REGIONS -
EU Strategy for Sustainable and Circular Textiles}",
  year={2022},
  url="{{https://eur-lex.europa.eu/legal-content/EN/TXT/?uri=celex:52022DC0141}}",
}

@misc{eu_dpp_textiles,
  author={{European Parliamentary Research Service}},
  title="{Digital product passport for the textile sector}",
  year={2024},
  url="{{https://www.europarl.europa.eu/RegData/etudes/STUD/2024/757808/EPRS_STU(2024)757808_EN.pdf}}",
}

@misc{ollama,
    title = {Ollama Website},
    author = {{Ollama}},
    howpublished = {\url{https://ollama.com/}},
    note = {Accessed: 2026-02-02}
}

@misc{yolov8,
    title = {you only look once Website},
    author = {{yolov8}},
    howpublished = {\url{https://yolov8.com/}},
    note = {Accessed: 2025-06-04}
}

@misc{sam,
    title = {segment anything Website},
    author = {{META}},
    howpublished = {\url{https://segment-anything.com/}},
    note = {Accessed: 2025-06-04}
}

@misc{llama32,
    title = {Llama 3.2: Revolutionizing edge AI and vision with open, customizable models},
    author = {{Meta Inc.}},
    howpublished = {\url{https://ai.meta.com/blog/llama-3-2-connect-2024-vision-edge-mobile-devices/}},
    note = {Accessed: 2025-04-06}
}

@misc{llama4,
    title = {The Llama 4 herd: The beginning of a new era of natively multimodal AI innovation},
    author = {{Meta Inc.}},
    howpublished = {\url{https://ai.meta.com/blog/llama-4-multimodal-intelligence/}},
    note = {Accessed: 2026-02-02}
}

@misc{gemma3,
    title = {Gemma 3: A family of lightweight models with multimodal understanding and unparalleled multilingual capabilities for more intelligent applications.},
    author = {{Google LLC}},
    howpublished = {\url{https://deepmind.google/models/gemma/gemma-3/}},
    note = {Accessed: 2026-02-02}
}

@misc{llava,
      title={Visual Instruction Tuning}, 
      author={Haotian Liu and Chunyuan Li and Qingyang Wu and Yong Jae Lee},
      year={2023},
      eprint={2304.08485},
      archivePrefix={arXiv},
      primaryClass={cs.CV},
      url={https://arxiv.org/abs/2304.08485}, 
}

@article{yao2024minicpm,
  title={MiniCPM-V: A GPT-4V Level MLLM on Your Phone},
  author={Yao, Yuan and Yu, Tianyu and Zhang, Ao and Wang, Chongyi and Cui, Junbo and Zhu, Hongji and Cai, Tianchi and Li, Haoyu and Zhao, Weilin and He, Zhihui and others},
  journal={arXiv preprint arXiv:2408.01800},
  year={2024}
}

@article{Qwen3VL,
      title={Qwen3-VL Technical Report}, 
      author={Shuai Bai and Yuxuan Cai and Ruizhe Chen and Keqin Chen and Xionghui Chen and Zesen Cheng and Lianghao Deng and Wei Ding and Chang Gao and Chunjiang Ge and Wenbin Ge and Zhifang Guo and Qidong Huang and Jie Huang and Fei Huang and Binyuan Hui and Shutong Jiang and Zhaohai Li and Mingsheng Li and Mei Li and Kaixin Li and Zicheng Lin and Junyang Lin and Xuejing Liu and Jiawei Liu and Chenglong Liu and Yang Liu and Dayiheng Liu and Shixuan Liu and Dunjie Lu and Ruilin Luo and Chenxu Lv and Rui Men and Lingchen Meng and Xuancheng Ren and Xingzhang Ren and Sibo Song and Yuchong Sun and Jun Tang and Jianhong Tu and Jianqiang Wan and Peng Wang and Pengfei Wang and Qiuyue Wang and Yuxuan Wang and Tianbao Xie and Yiheng Xu and Haiyang Xu and Jin Xu and Zhibo Yang and Mingkun Yang and Jianxin Yang and An Yang and Bowen Yu and Fei Zhang and Hang Zhang and Xi Zhang and Bo Zheng and Humen Zhong and Jingren Zhou and Fan Zhou and Jing Zhou and Yuanzhi Zhu and Ke Zhu},
	  journal={arXiv preprint arXiv:2511.21631},
      year={2025}
}

@article{Ergun2025_textile_vlm,
title = {Towards Automated Handling and Sorting of Garments combining Visual Language Models and Convolutional Neural Networks},
journal = {Proceedings of the Austrian Robotics Workshop 2025},
pages = {25-30},
year = {2025},
issn = {3061-0710},
url = {https://www.fh-salzburg.ac.at/fileadmin/fhs_daten/departments/information-technologies/documents/ARW2025_Proceedings_final_kl.pdf},
author = {Serkan Ergun and Tobias Mitterer and Hubert Zangl}
}

@InProceedings{radford2021_CLIP,
  title = 	 {Learning Transferable Visual Models From Natural Language Supervision},
  author =       {Radford, Alec and Kim, Jong Wook and Hallacy, Chris and Ramesh, Aditya and Goh, Gabriel and Agarwal, Sandhini and Sastry, Girish and Askell, Amanda and Mishkin, Pamela and Clark, Jack and Krueger, Gretchen and Sutskever, Ilya},
  booktitle = 	 {Proceedings of the 38th International Conference on Machine Learning},
  pages = 	 {8748--8763},
  year = 	 {2021},
  editor = 	 {Meila, Marina and Zhang, Tong},
  volume = 	 {139},
  series = 	 {Proceedings of Machine Learning Research},
  publisher =    {PMLR},
  url = 	 {https://proceedings.mlr.press/v139/radford21a.html}
}

@ARTICLE{kawaharazuka2025_VLA_review,
  author={Kawaharazuka, Kento and Oh, Jihoon and Yamada, Jun and Posner, Ingmar and Zhu, Yuke},
  journal={IEEE Access}, 
  title={Vision-Language-Action Models for Robotics: A Review Towards Real-World Applications}, 
  year={2025},
  volume={13},
  number={},
  pages={162467-162504},
  doi={10.1109/ACCESS.2025.3609980}
}

@article{kim24openvla,
    title={OpenVLA: An Open-Source Vision-Language-Action Model},
    author={{Moo Jin} Kim and Karl Pertsch and Siddharth Karamcheti and Ted Xiao and Ashwin Balakrishna and Suraj Nair and Rafael Rafailov and Ethan Foster and Grace Lam and Pannag Sanketi and Quan Vuong and Thomas Kollar and Benjamin Burchfiel and Russ Tedrake and Dorsa Sadigh and Sergey Levine and Percy Liang and Chelsea Finn},
    journal = {arXiv preprint arXiv:2406.09246},
    year={2024}
}

@misc{nvidia2025gr00tn1openfoundation,
      title={GR00T N1: An Open Foundation Model for Generalist Humanoid Robots}, 
      author={NVIDIA and : and Johan Bjorck and Fernando Castañeda and Nikita Cherniadev and Xingye Da and Runyu Ding and Linxi "Jim" Fan and Yu Fang and Dieter Fox and Fengyuan Hu and Spencer Huang and Joel Jang and Zhenyu Jiang and Jan Kautz and Kaushil Kundalia and Lawrence Lao and Zhiqi Li and Zongyu Lin and Kevin Lin and Guilin Liu and Edith Llontop and Loic Magne and Ajay Mandlekar and Avnish Narayan and Soroush Nasiriany and Scott Reed and You Liang Tan and Guanzhi Wang and Zu Wang and Jing Wang and Qi Wang and Jiannan Xiang and Yuqi Xie and Yinzhen Xu and Zhenjia Xu and Seonghyeon Ye and Zhiding Yu and Ao Zhang and Hao Zhang and Yizhou Zhao and Ruijie Zheng and Yuke Zhu},
      year={2025},
      eprint={2503.14734},
      archivePrefix={arXiv},
      primaryClass={cs.RO},
      url={https://arxiv.org/abs/2503.14734}, 
}

@article{Mazumder2023,
title = {Towards next generation digital twin in robotics: Trends, scopes, challenges, and future},
journal = {Heliyon},
volume = {9},
number = {2},
pages = {e13359},
year = {2023},
issn = {2405-8440},
doi = {https://doi.org/10.1016/j.heliyon.2023.e13359},
url = {https://www.sciencedirect.com/science/article/pii/S2405844023005662},
author = {A. Mazumder and M.F. Sahed and Z. Tasneem and P. Das and F.R. Badal and M.F. Ali and M.H. Ahamed and S.H. Abhi and S.K. Sarker and S.K. Das and M.M. Hasan and M.M. Islam and M.R. Islam}
}

@INPROCEEDINGS{Schoenberger2016_object_reconstruction,
  author={Schönberger, Johannes L. and Frahm, Jan-Michael},
  booktitle={2016 IEEE Conference on Computer Vision and Pattern Recognition (CVPR)}, 
  title={Structure-from-Motion Revisited}, 
  year={2016},
  volume={},
  number={},
  pages={4104-4113},
  doi={10.1109/CVPR.2016.445}
  }

@article{Varga2017_arrowhead_eclipse,
author = {Varga, Pal and Blomstedt, Fredrik and Ferreira, Luis Lino and Eliasson, Jens and Johansson, Mats and Delsing, Jerker and Martnez de Soria, Iker},
title = {Making system of systems interoperable The core components of the arrowhead framework},
year = {2017},
issue_date = {March 2017},
publisher = {Academic Press Ltd.},
address = {GBR},
volume = {81},
number = {C},
issn = {1084-8045},
journal = {J. Netw. Comput. Appl.},
month = mar,
pages = {85–95},
numpages = {11}
}

@INPROCEEDINGS{Herguedas2019_deformable_survey,
  author={Herguedas, Rafael and López-Nicolás, Gonzalo and Aragüés, Rosario and Sagüés, Carlos},
  booktitle={2019 24th IEEE International Conference on Emerging Technologies and Factory Automation (ETFA)}, 
  title={Survey on multi-robot manipulation of deformable objects}, 
  year={2019},
  volume={},
  number={},
  pages={977-984},
  doi={10.1109/ETFA.2019.8868987}
}

@article{Billard2019_robot_manipulation_trends,
author = {Aude Billard  and Danica Kragic },
title = {Trends and challenges in robot manipulation},
journal = {Science},
volume = {364},
number = {6446},
pages = {eaat8414},
year = {2019},
doi = {10.1126/science.aat8414},
URL = {https://www.science.org/doi/abs/10.1126/science.aat8414},
eprint = {https://www.science.org/doi/pdf/10.1126/science.aat8414}
}

@ARTICLE{EGAD,
 author={Morrison, Douglas and Corke, Peter and Leitner, Jürgen},
journal={IEEE Robot. Autom. Lett.}, 
 title={EGAD! An Evolved Grasping Analysis Dataset for Diversity and Reproducibility in Robotic Manipulation}, 
year={2020},
volume={5},
number={3},
pages={4368-4375},
doi={10.1109/LRA.2020.2992195}}

@ARTICLE{CapTac,
  author={Ergun, Serkan and Mishra, Rishabh B. and Ananadan, Narendiran and Mitterer, Tobias and Mattoli, Virgilio and Zangl, Hubert},
  journal={IEEE Robotics and Automation Letters}, 
  title={CapTac: Robust Capacitive Sensing for Distributed Force Mapping in Parallel Robotic Grasping}, 
  year={2026},
  volume={11},
  number={3},
  pages={3668-3675},
  doi={10.1109/LRA.2026.3662613}}
}


\end{document}